\documentclass{article}

\usepackage{arxiv}

\usepackage[utf8]{inputenc} % allow utf-8 input
\usepackage[T1]{fontenc}    % use 8-bit T1 fonts
\usepackage{hyperref}       % hyperlinks
\usepackage{url}            % simple URL typesetting
\usepackage{booktabs}       % professional-quality tables
\usepackage{amsfonts}       % blackboard math symbols
\usepackage{nicefrac}       % compact symbols for 1/2, etc.
\usepackage{microtype}      % microtypography
\usepackage{graphicx}
\usepackage{doi}

\usepackage{stackengine}
\usepackage{subcaption}

\usepackage{multirow}
\usepackage{wrapfig}
\usepackage{mysymbols}

\begin{document}
\title{HomeRobot Open Vocabulary Mobile Manipulation Challenge 2023 Participant Report}
\author{{Volodymyr Kuzma} \\
        \And
        {Vladyslav Humennyy} \\
        \And
        {Ruslan Partsey} \\
}
\date{Ukrainian Catholic University\\\{volodymyr.kuzma, vladyslav.humennyi, partsey\}@ucu.edu.ua}
% UCU
% (mail, mail, mail)@ucu.edu.ua

\maketitle

\begin{abstract}
We report an improvements to NeurIPS 2023 HomeRobot: Open Vocabulary Mobile Manipulation (OVMM) Challenge reinforcement learning baseline. More specifically, we propose more accurate semantic segmentation module, along with better place skill policy, and high-level heuristic that outperforms the baseline by 2.4\% of overall success rate (sevenfold improvement) and 8.2\% of partial success rate (1.75 times improvement) on Test Standard split of the challenge dataset. With aforementioned enhancements incorporated our agent scored 3rd place in the challenge on both simulation and real-world stages.
\end{abstract}

\keywords{OVMM Challenge, YOLOv8, MobileSAM, Detic, Semantic Segmentation.}

% \begin{figure}[!h]
%     \centering
%     \includegraphics[width=0.8\linewidth]{images/GeneralPipeline.pdf}
%     \caption{RL-based OVMM agent pipeline. \textbf{Left:} Segmentation pipeline. YOLOv8~\cite{yolov8_ultralytics} and MobileSAM~\cite{mobile_sam} are used in addition to original Detic~\cite{zhou2022detecting} segmentation. \textbf{Center:} Low-level RL-based policy. Obtained segmentation mask, along with RGB, depth frames, and other measurements, like \textit{GPS~and~Compass}, are used to sample a discrete (or continuous in case of Gaze and Place skills) action. \textbf{Right:} High-level heuristic. Skills are called in sequence with conditional loop in case if the pick is unsuccessful.}
%     \label{fig:gen_architect}
% \end{figure}

\section{Introduction}

Nowadays, robotics is rapidly developing, making it more and more used across various domains. However, it is still non-trivial task for robots to navigate and interact in human spaces, such as homes. Therefore, significant resources are used to improve performance of embodied agents in domestic conditions across wide range of tasks.

\subsection{Motivation}
Yenamandra~\etal~\cite{homerobot} revealed that open-vocabulary mobile manipulation task (fundamental for embodied agents) was not standardized and described, making reproduction, comparison, and, therefore, improvement of existing methods nearly impossible. According to their study, current heuristic (composed of heuristic planners) and reinforcement learning (composed of low-level RL-trained skills) approaches could achieve at most 11.6\% success rate with ground truth semantic segmentation and only at most 0.8\% success rate with open-vocabulary segmentation mask computed online by Detic~\cite{zhou2022detecting}. These findings show that OVMM task is still an open frontier for the Embodied AI (EAI) research.

In this work, we report changes of the RL baseline proposed in the HomeRobot paper~\cite{homerobot}, improving semantic understanding of the agent by utilizing retrained YOLOv8~\cite{yolov8_ultralytics} object detection model and MobileSAM~\cite{mobile_sam} segmentation model. We also analyze the performance of baseline's place skill to improve training reward function and retrain the corresponding policy.

Our agent reached the final real-world evaluation stage and was ranked third on NeurIPS 2023 HomeRobot: OVMM Challenge~\cite{homerobotovmmchallenge2023}. Nevertheless, even with all the enhancements our approach is still far from solving the task, achieving only 2.8\% overall success rate on the Test Standard split of the challenge.
% Description of robot and its capabilities?

\subsection{Report Structure}
The report is structured as follows: Section~\ref{section:task_description} gives a detailed description of the OVMM task, Section~\ref{section:OVMMCHallenge} introduces the OVMM Challenge, in Section~\ref{section:explor_analysis} we identify the baseline's limitations, Section~\ref{section:agent} describes proposed agent improvements, Section~\ref{section:place} includes analysis and training parameters for place skill, Section~\ref{section:results} shows our agent's results in local experiments (local simulation in Habitat~\cite{savva2019habitat, szot2021habitat} environment) and on challenge leaderboard, Section~\ref{section:possible_improvements} lists areas for possible improvements, and in Section~\ref{section:conclusions}, we make concluding remarks.

 % Finally, we make conclusive remarks in Chapter 5.

\section{OVMM Task}
\label{section:task_description}
OVMM is an EAI task in which an agent has to navigate in an unknown environment, find specified object, and move (pick-navigate-place) this object to a specified target receptacle. The agent has to have the ability to recognize and handle any objects (even never-before-seen categories during training) described in a text prompt (open-vocabulary agent). At the same time, such condition is not placed on receptacles: all categories are seen in the training set. That is an essential task for home robots, as they have to be able to understand and execute any command in the context of a house.

\begin{wrapfigure}[23]{r}{0.35\textwidth}
  \centering
  \begin{minipage}{0.28\textwidth}
    \centering
    \includegraphics[width=\textwidth]{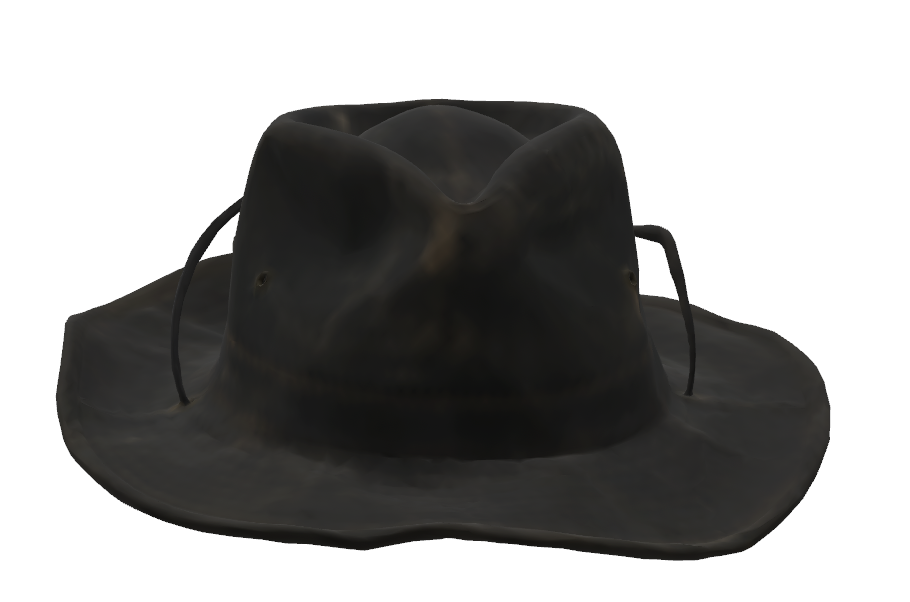}
  \end{minipage}
  \hfill
  \begin{minipage}{0.28\textwidth}
    \centering
    \includegraphics[width=\textwidth]{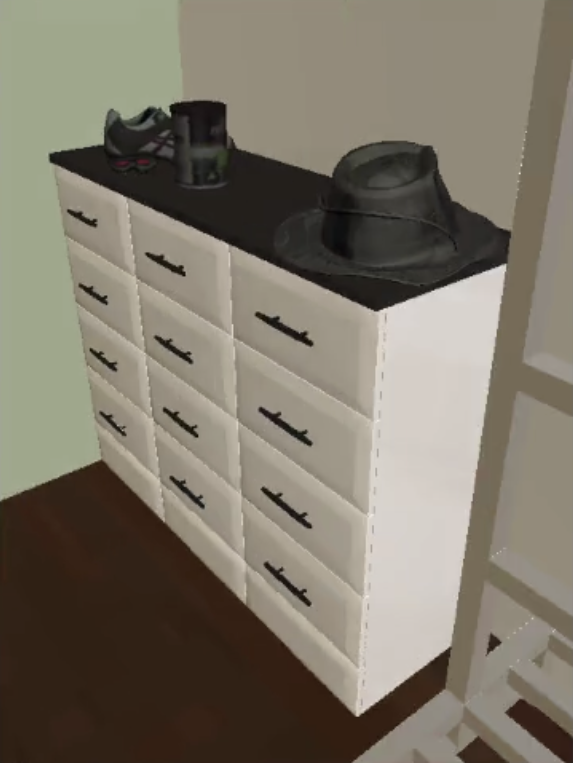}
  \end{minipage}
  \caption{Example of goal object (hat) on start receptacle (cabinet).}
  \label{fig:hat_example}
\end{wrapfigure}

At the start of an episode, the agent is provided with a prompt in the following form: "Move ($object$) from the ($start\_receptacle$) to the ($goal\_receptacle$)." $Object$ is a movable entity that can be grasped (see Figure~\ref{fig:hat_example}). Receptacles ($\{start,goal\}\_receptacle$) are different types of furniture that have surfaces suitable for placement. Receptacles can vary in height, size, and the presence of additional non-surface parts (e.g., the back of a chair). Besides, there may be several instances of the objects and receptacles in scene. However, it is important that the $object$ is picked from the specified $start\_receptacle$ and placed on any $goal\_receptacle$.

The task is considered successful if any $object$ instance is picked up from any $start\_receptacle$ instance and then placed on any $goal\_receptacle$ instance. Besides contact with the receptacle, $object$ must have a stable position after placement.

For evaluation and execution purposes, the task is divided into four distinct subtasks, which are considered for the agent's partial success:

\begin{enumerate}
    \item \subtask{Navigation to object}
    \item \subtask{Pick (gaze)}
    \item \subtask{Navigation to receptacle}
    \item \subtask{Place}
\end{enumerate}

These subtasks are executed sequentially, and successful completion of one requires the completion of all preceding subtasks. The success of the final subtask, \subtask{Place}, counts as the overall success of the task.

\subsection{Navigation to object}

\subtask{Navigation to object} is the initial subtask, which starts navigation within the environment and aims to find and reach the $object$ placed on $start\_receptacle$. The agent searches for the $object$ by executing navigation actions (\eg go forward, turn left/right in case of discrete action space). To successfully complete this subtask, the agent must come close to  $object$ and keep it in sight.

\subsection{Pick (Gaze)}

The \subtask{pick} subtask is different in simulation and real-world. In the simulation, the task is limited to gazing at the goal $object$ without the requirement to manually pick it up. This implementation is due to limitations in recreating manipulator-object interactions within the simulation environment. Therefore, in the simulation, agent only has to gaze directly at the $object$ and stand in close proximity to initiate the Snap action, which then teleports the $object$ into the robot manipulator. However, in real-world scenarios, this part of the task is executed entirely through the physical picking of the $object$ using the robot manipulator.

\subsection{Navigation to receptacle}

After picking up the $object$, the agent must find any $goal\_receptacle$ in the environment and approach it. Success is counted only if the agent is close to the $goal\_receptacle$ instances.

\subsection{Place}

In the final stage of the task, the agent must place the $object$ on the $goal\_receptacle$. To achieve this, the agent can use both manipulation and navigation skills, which add degrees of freedom (DoF) to the final part of the task. Additionally, successful placement requires that the $object$ is in a stable position on the $goal\_receptacle$. In the simulation, this is determined by examining the $object$'s speed; if it does not exceed a certain threshold, the $object$ is considered to be in a stable position.

\section{OVMM Challenge}
\label{section:OVMMCHallenge}

The challenge~\cite{homerobotovmmchallenge2023} was divided into two phase: the virtual and the real-world. Based on the results of the first part, top-3 teams were selected to proceeded to the final stage of the competition, at which the organizers tested the agents using real-world facilities.

The final results of the challenge were determined based on the real-world phase that the organizers conducted.

\subsection{Virtual Phase}

% For the challenge, a specified repository was created\footnote{https://github.com/facebookresearch/home-robot}, which contained all required assets and scripts for the execution of the virtual phase of the challenge. For comfort of participants, an easy interface was realised for the interactions of robots with an environment. Besides, available RL and heuristic baseline agents had pretty low success rates, but were helpful initial agents upon which to build.

The virtual phase of challenge  was run from 1 July 2023 to 20 October 2023. During that time, participants were developing their agents in the virtual environment Habitat~\cite{savva2019habitat, szot2021habitat}, in which OVMM task and basic robot logic were implemented. The scenes were used from Habitat Synthetic Scene Dataset~\cite{khanna2023habitat}.

Using those instruments, participants had to solve OVMM tasks described in detail in Section \ref{section:task_description}. Environment action space could be divided into two parts: navigation and manipulation. Navigation skills (\skill{navigation to object}, \skill{navigation to receptacle})\footnote{\skill{Skills} are separate policies that were meant to be used by agent for certain \subtask{subtask}. More about them in Subsection \ref{subsection:skill}} were those that let you move in the environment effectively. They were discrete, with three primary actions - go forward, turn left, and turn right. On the other hand, manipulation skills (\skill{pick}, \skill{place}) gave access to both the robot arm and stand, which was helpful for accurate interactions with objects for picking or placing. This part of the task used continuous action space, which let an agent make even the slightest move to correct its base/arm positions.

Agents were evaluated on EvalAI~\cite{evalai}, a platform designed for evaluating and comparing ML models. Participants submitted their Docker containers with executable agents to the platform, where they got scores on certain dataset splits. Results at Test Challenge Split were used to determine top-3 teams, which then proceeded to the real-world phase.

\subsection{Real-World Phase}

During real-world phase, a dedicated scene was constructed. OVMM task was carried out there with use of Hello Robot Stretch robot~\cite{stretch}. 

% To test the top three agents in real-world conditions, a dedicated facility will be set up equipped with a Hello Robot Stretch, the same robot model used in the virtual training phase. This allows us to directly evaluate agents' ability to transfer their simulated skills to an actual robot and environment.

\section{Exploratory Analysis}
\label{section:explor_analysis}
Our initial step was the analysis of NeurIPS 2023 HomeRobot: Open Vocabulary Mobile Manipulation Challenge~\cite{homerobotovmmchallenge2023}, as we were looking for more information about current agent, environment and problems that we could encounter during the challenge. We identified two main components that had significant impact on the agent's success: perception and \skill{place} skill. Thus, we decided to focus on improving these two components.

\subsection{Perception Impact}
\label{subsection:perception_impact}
For both heuristic and RL agent types, perception played a crucial role in algorithm performance, substantially influencing all subtasks and overall success. All algorithms employed semantic masks to recognize the required objects and receptacles within the environment. These masks were obtained using perception modules, and the authors of the paper~\cite{homerobot} compared two methods for computing them: ground truth and Detic (see Table \ref{table:gt_detic_comparison}).

\begin{table}[htbp]
\centering
\caption{Comparison Table for Ground Truth and Detic RL Agent (Absolute Percent)\cite{homerobot}}
\begin{tabular}{lcccccccc}
\toprule
\textbf{Perception} & \multicolumn{3}{c}{\textbf{Partial Success Rates}} & \textbf{Overall} & \textbf{Partial} \\
 & \subtask{FindObj\footnotemark} & \subtask{Pick} & \subtask{FindRec} & \textbf{Success Rate} & \textbf{Success Metric} \\
\midrule
Ground Truth & 55.7 & 50.2 & 35.2 & 11.6 & 38.2 \\
\midrule
Detic & 19.8 & 11.8 & 6.3 & 0.8 & 9.7 \\
\bottomrule
\end{tabular}\\
\label{table:gt_detic_comparison}
\end{table}

Direct comparison revealed a stark contrast: transitioning to the Detic perception module resulted in a four-fold decrease in partial success rate and a ten-fold decrease in overall success rate. This led to the conclusion that the initial agent had significant potential for improvement through a perception upgrade alone. Moreover, this improvement positively influenced all of the agent's skills and ultimately had a greater impact on the overall task performance than upgrading a single skill.
\newpage
\subsection{\skill{Place} Skill Bottleneck}
\label{subsection:place_impact}

\begin{wrapfigure}{r}{0.5\textwidth}
    \centering
    \includegraphics[width=0.45\textwidth]{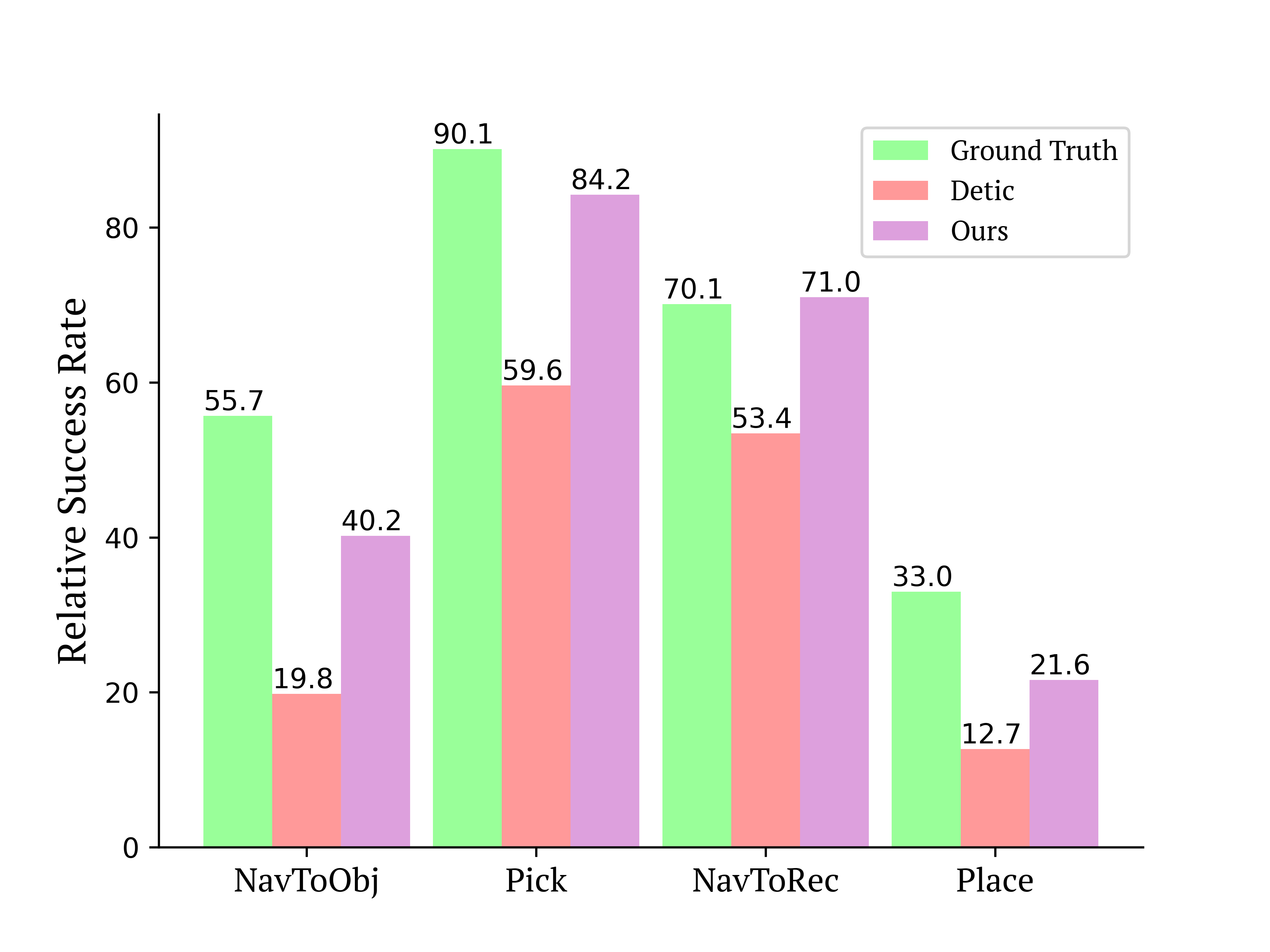}
    \caption{Skill relative success rates.}
    \label{fig:baseline_res}
\end{wrapfigure}

Beyond the overall impact of perception on the entire task, we also investigated the changes in performance for individual skills with different perception modules (Figure~\ref{fig:baseline_res}). The \skill{place} skill stood out as the weakest performer, with the lowest relative success rate under both perception conditions (33.0\% and 12.7\%). Additionally, it, along with the \skill{navigation to object} skill, experienced a significant drop in success (approximately 3 times) when shifting from ground truth to Detic (for \skill{navigation to object} success dropped from 55.7\% to 19.8\%). This decrease was probably caused by the problems Detic faces in recognizing small objects. Notably, the \skill{navigation to receptacle} skill exhibited the smallest relative drop (from 70.1\% to 53.4\%), suggesting that recognizing large furniture was not a serious hurdle for Detic.

In contrast, the root cause of the drop in \subtask{place} task success is not obvious. Thus, we decided to examine it more closely during our experiments.

\section{Our agent}
\label{section:agent}

\subsection{High-level Heuristic}
\label{subsection:skill}
\begin{figure}[htbp]
    \centering   \includegraphics[width=1\linewidth]{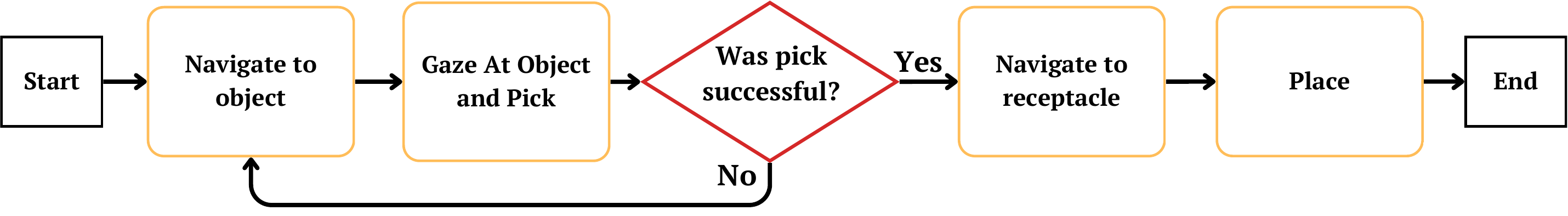}
    \caption{Original high-level policy.}
    \label{fig:highlevel_policy}
\end{figure}
RL baseline, on which our approach was based, is divided into four skills that correspond to the separate parts of the task: a skill for \skill{navigation to object}; a skill for \skill{pick} that is called \skill{gaze} and actually performs only the preparation for pick, as in simulation the process of grasping is automatic; a skill for \skill{navigation to receptacle}, and one for \skill{placing} of the object. To control and switch skills a high-level policy is used. In case of the baseline the skills are simply called one by one in a row: when previous skill calls Stop action, it is terminated and switched to next one. 
\footnotetext{\subtask{Find object} is named \subtask{navigation to object} in our work. Similarly, \subtask{find receptacle} is named \subtask{navigation to receptacle}.}As the next phases can not be successful unless all previous ones are, it is useful to check previous one's success. For navigation skills (\skill{navigation to object}, \skill{navigation to receptacle}) and \skill{place} skill, it is a non-trivial task to distinguish successful performance from the unsuccessful one. On the other hand, for \skill{pick} skill there is an accessible sensor measurement that directly shows whether the item was actually picked. Considering this, we added a conditional loop over first two skills that was performed until the pick was successful.\footnote{The \skill{navigation to object} skill was also inside the loop because of the constant wandering far from the object during unsuccessful pick attempts, making restart of only \skill{Gaze} skill mostly fruitless.}

The change of high-level heuristic is better described in Figure~\ref{fig:highlevel_policy}.

\subsection{Low-Level Policies}

\begin{figure}[t]
    \centering
    \includegraphics[width=0.5\linewidth]{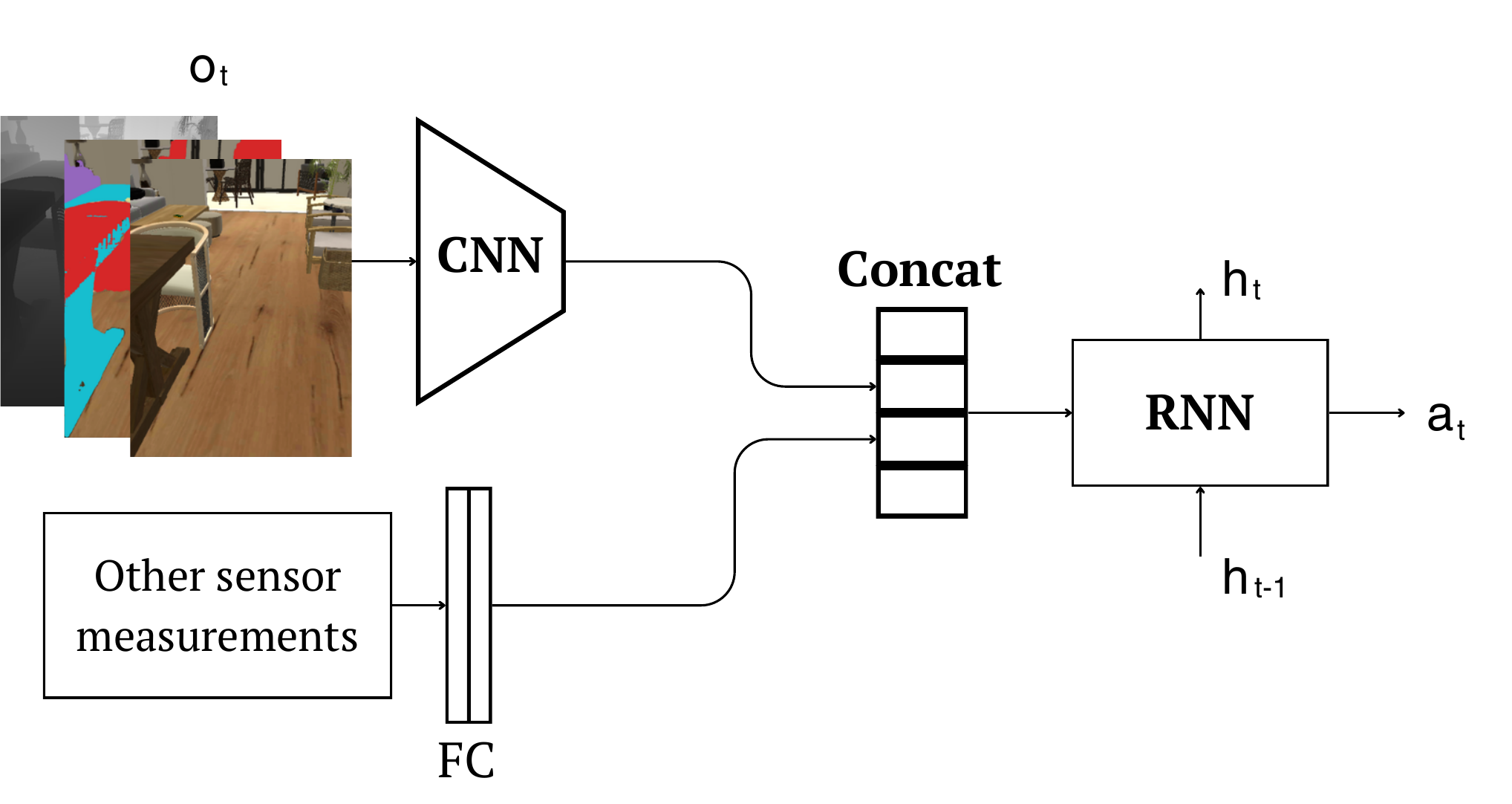}
    \caption{Architecture of a low-level policy.}
    \label{fig:lowelevel_policy}
\end{figure}

In the baseline, low-level policies, responsible for all skills, are neural networks where RGB, depth, and semantic segmentation mask are passed through a CNN (ResNet~\cite{he2015deep} in our case). Then they are concatenated with features obtained from other measurements (like GPS and Compass) passed through fully connected layers, and put into an LSTM~\cite{hochreiter1997long} block, along with previous step state $h_{t-1}$ to get new state $h_t$ and distribution, from which action will be sampled (discrete if skill is navigational and continuous if skill requires precise movements like \skill{gaze} or \skill{place}). The architecture is shown in Figure~\ref{fig:lowelevel_policy}.

We did not change the architecture of the policies, working only on retraining of the existing ones (more in section \ref{section:place}).

\subsection{Perception Module}
\begin{figure}[b]
    \centering
    \includegraphics[width=0.7\linewidth]{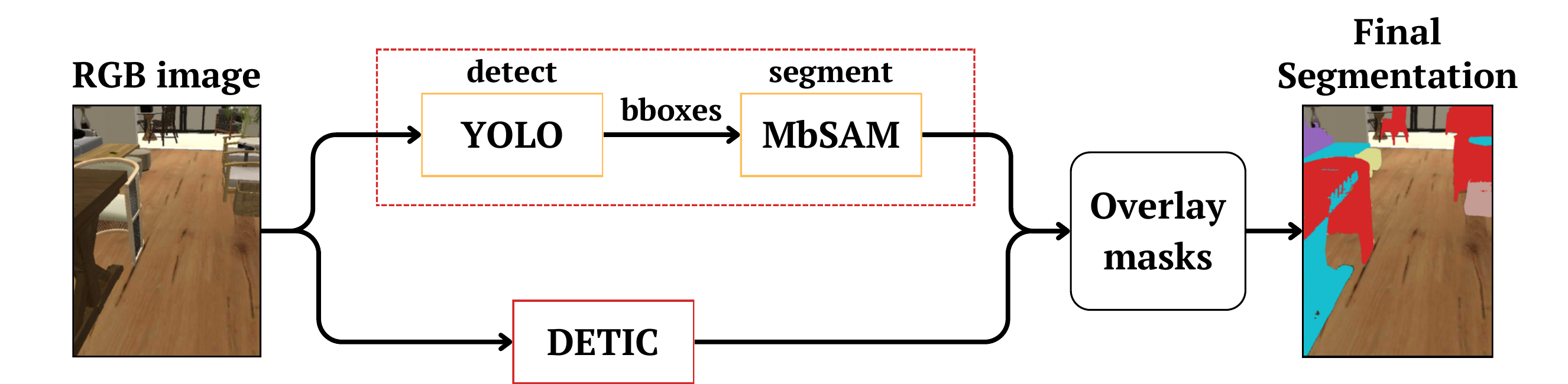}
    \caption{Semantic segmentation pipeline.}
    \label{fig:segmentation}
\end{figure}

As shown in Subsection \ref{subsection:perception_impact}, the baseline, trained with ground truth semantic segmentation, shows lower performance when computing segmentation mask online (no ground truth). Our approach enhanced existing Detic~\cite{zhou2022detecting} perception, utilizing MobileSAM~\cite{mobile_sam} model for segmentation, prompted with bounding boxes from YOLOv8~\cite{yolov8_ultralytics} detection model, retrained on a dataset of 50000 images from scenes with objects and receptacles on them. Obtained segmentation masks for each class were overlaid, putting the task-important objects (goal object, start receptacle, and end receptacle classes) on top of the resulting mask. Detic's segmentation mask was merged with the previous one in places where YOLO-SAM module failed to detect any objects, and for regions where Detic detected goal object. This way, the final semantic segmentation module preserved strengths of the open-vocabulary Detic while adding more task-specific YOLO-SAM, good at distinguishing furniture types and small objects. Perception module scheme is visualized in Figure~\ref{fig:segmentation}.

\section{Place Improvement}
\label{section:place}

\subsection{Baseline Performance Analysis}
As concluded in subsection \ref{subsection:place_impact}, \skill{place} skill significantly impacts overall success rate of the baseline (also fail in place results in 0 overall success even though all previous skills succeded). To better understand the reasons behind some failed episodes, we analyzed the agent's performance in 32 cases where the final phase was reached. Here are all the observed causes of failure:
\begin{itemize}
    \item unstable place - 25.0\%;
    \item missed receptacle - 21.9\%;
    \item not annotated receptacle - 15.6\%;
    \item camera overlap with manipulator - 15.6\%;
    \item receptacle misdetection - 9.4\%;
    \item did not start place skill - 6.3\%;
    \item uncertain - 6.3\%.
\end{itemize}

Though we were not able to address all of these cases in our work, we hope that this analysis can help future efforts to improve the baseline. At the same time, other failures could be resolved via adjustment of training reward of the skill (by penalizing the actions that lead to the errors or remaking the sparse reward into continuous for faster learning). Therefore, we started experimenting with reward changes.

\subsection{Reward Engineering}

The original baseline's \skill{place} skill was trained on a sparse reward: the agent got +5 for drop on the receptacle surface when object came into contact with it and +1 for each timestep when the object remains on the surface. A penalty of -1 was given when the object was dropped and did not come into contact with the surface~\cite{homerobot}. Analyzing data from the previous part, we highlighted important features to make reward better:
\begin{itemize}
    \item strictly penalizes wandering very far from the starting position to avoid movement away from the goal (common due to noisy semantic segmentation), as it is known that the goal should be close to the agent at the beginning;
    \item rewards approaching the goal as close as possible;
    \item rewards keeping the goal object in sight;
    \item penalizes blocking camera view with manipulator;
    \item penalizes waiting when in the final position to place (very close to the object and looking at it).
\end{itemize}

To implement the features we added continuous fixed\footnote{This means that for each episode there is the same total possible reward for these actions achieved when agent reaches some threshold of distance or presence of goal in camera (measured as percentage of pixels belonging to goal)} rewards for approaching the goal and keeping it in the camera view.
For penalties, we used constant reward reductions when reaching certain thresholds each timestep (for example, -3 each step when farther than 1.5m from start). 

Final training reward parameters were:
\begin{itemize}
    \item reward for place when object came into contact with surface is 70;
    \item reward for object being in contact with surface each step (per step) is 25;
    \item total distance reward is 40;
    \item minimal distance to goal threshold (when reward for distance reduction is no more given) is 0.2m;
    \item total view reward is 30;
    \item goal visibility threshold (after this threshold reward for bigger percentage of goal receptacle in screen space is no more given) is 30\% ;
    \item camera blocking penalty (per step) is -5;
    \item wandering far from start penalty (per step) is -5.
\end{itemize}

After selecting reward and threshold values, we performed final training of place skill, starting from the weights of original baseline.

\subsection{Training Details}

We performed experiments on two machines: first setup with 2 RTX3090 GPUs with 24Gb memory was used for parameter adjustment, second machine with 4 A6000 GPUs with 48Gb memory was used for final training. We ran 32 environments per machine. The learning rate was set to $10^{-7}$ for better adaptation of original weights. All the other parameters remained the same as in original implementation.

\section{Results}
\label{section:results}

\subsection{Our Results}

After implementation of each feature in our agent, we evaluated it on given validation split to gain information about improvements of agent at task. They are showed in Table \ref{table:res_table}.

\begin{table}[h!]
\centering
\caption{Skill and Task Success Rate.}
\label{table:res_table}
\scalebox{.9}{\begin{tabular}{lccccccc}
\hline
\multirow{2}{*}{Perception} & \multicolumn{4}{c}{Skill relative success rate} & 
\multirow{2}{*}{\begin{tabular}{@{}c@{}}
Overall \\
Success Rate
\end{tabular}} & 
\multirow{2}{*}{\begin{tabular}{@{}c@{}}
Partial \\
Success Rate
\end{tabular}} \\ \cline{2-5}
 & \subtask{NavToObj} & \subtask{Pick} & \subtask{NavToRec} & \subtask{Place} & \\ \hline 
Detic Baseline & 19.8 & 59.6 & 53.3 & 12.5 & 0.8 & 9.7 &  \\
YOLO-SAM Baseline & 22.75 & 57.1 & 48.1 & 0 & 0 \brackets{(-0.8)} & 10.5 \brackets{(+0.8)} &  \\
YOLO-SAM Baseline + Detic & 38.4 & \textbf{85.4} & \textbf{77.1} & 15.8 & 4 \brackets{(+3.2)} & 25.1 \brackets{(+15.4)} &  \\
YOLO-SAM Fine-tuned + Detic & \textbf{40.2} & 84.2 & 71 & \textbf{21.6} & \textbf{5.2} \brackets{(+4.4)} & \textbf{25.8} \brackets{(+16.1)} &  \\ \hline
\end{tabular}
}
\end{table}

While introducing pure YOLO-SAM perception did not lead to an overall success rate increase, it did provide a boost in partial success rate. However, the most significant improvements came with the combined use of YOLO-SAM and Detic perception. In comparison to pure Detic, overall success rate improved fivefold, while the partial success rate improved 2.5 times. We can see that even when the relative success rate of the \skill{place} skill was almost the same, the overall success rate rose significantly, as we gained a lot in all of the previous tasks (see Figure~\ref{fig:baseline_res}), resulting in such a final result.

With the introduction of a new place checkpoint, we improved overall success rate, which resulted in a 1.2\% increase from our previous result.

\subsection{EvalAI Leaderboard}

In addition to evaluation result on our local machines we evaluated our agent on EvalAI~\cite{evalai}, where competition was hosted. There were two important splits: Test Standard and Test Challenge. First one was open and visible for all participants, while the second one was closed and was used to determine top-3 teams of the challenge.

As of \today, we are ranked first at Test Standard split, with 0.8 and 0.5 percents lead at overall success and partial success respectively (see Table \ref{table:evalai_leaderboard}).

\begin{table}[htbp]
\centering
\caption{EvalAI Challenge Test Standard Results}
\begin{tabular}{lcccccccc}
\toprule
\textbf{Rank} & \textbf{Participant Team} & \textbf{Overall Success} & \textbf{Partial Success} \\
\midrule
1 & KuzHum & \textbf{0.028} & \textbf{0.191} \\
2&UniTeam&0.020&0.186 \\
3&PieSquare&0.020&0.111 \\
4&VCEI&0.012&0.123 \\
5&PoorStandard (DJI)&0.008&0.110 \\
6&scale\_robotics&0.004&0.107 \\
\bottomrule
\end{tabular}
\label{table:evalai_leaderboard}
\end{table}

In the Test Challenge split, we got into top-3 teams, but without any specific information about our result at that split.

\subsection{Sim-to-Real Transfer}

\begin{figure}[b]
    \centering
    \includegraphics[width=0.5\linewidth]{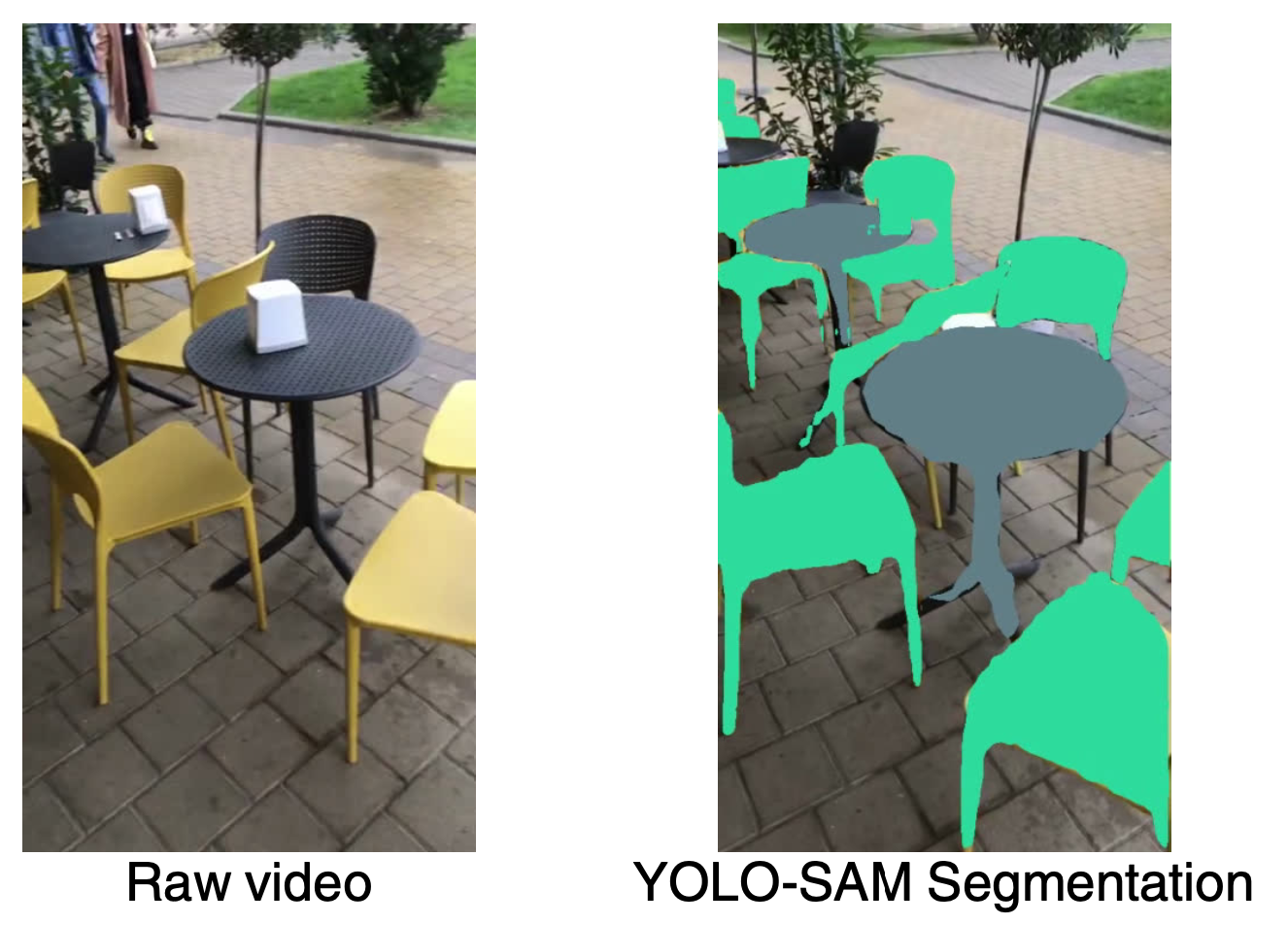}
    \caption{Real-life perception results.}
    \label{fig:real_world_viz}
\end{figure}

A very important part of the challenge~\cite{homerobotovmmchallenge2023} was evaluation of robot in real world. We were curious about performance of our agent outside the simulation (similarly to the work done by Anderson, \etal~\cite{pmlr-v155-anderson21a}). As we lacked any physical robot with manipulators, we decided to perform test of our perception module, that we were working on. Thus, we tested how perception, trained in simulations at virtual assets, performed in real world. After creation of several segmentation videos of both indoor and outdoors receptacles we saw, that perception module easily recognizes receptacles as chairs, tables (see Figure~\ref{fig:real_world_viz}). At similar courses with different number of objects and different illumination, perception gave similar results.

\section{Future Work}
\label{section:possible_improvements}
As task of OVMM is far from being ‘solved’, we share our insights and ideas for future work below. 

% \subsection{Current Limitations}
% May be 

\subsection{Object Tracker}

During navigation and manipulations, the perception module calls inference at each frame without including information from previous steps. Thus, during inspection of the environment, we have seen that on consecutive frames semantic masks of the same object may disappear (which does not happen during training with ground truth semantic segmentation). This leads to unpredictable behavior of the agent, which can get lost in the environment.

To prevent this problem, we can implement an object tracker into the perception module. It would use information from both the current and previous frames, which would prevent disappearing of an object from frame to frame. As the agent is trained with use of ground truth, which has correct segmentation at all frames, we suggest that this may improve its behavior, as with better perception evaluation would be more similar to training conditions.

\subsection{Policy and Skills}
% Vladyslav

Besides technical improvements to the agent's perception modules, its performance also relies on its logic and actions. Therefore, improving the agent's policies to get more efficient behavior throughout the episode is crucial. We have outlined two main components in this part of the agent: a high-level policy which controls calls to skills, and each separate skill that performs a specific part of the task.

\subsubsection{High-Level Policy}

Throughout the entire episode, the agent is controlled by a high-level policy. This policy receives information about the agent's state in the environment, including its position and interactions with goal objects. Based on this information, the policy chooses which skill to execute at the moment.

The current version of the policy is quite simple: it sequentially calls skills in their logical order. Additionally, during our development of the agent, we also implemented an improvement to the \skill{gaze} skill (see Figure~\ref{fig:highlevel_policy}). If the agent does not successfully complete this subtask (does not have an object in its arm), the skill is executed repeatedly. Even this basic policy change improved our success rates. However, this policy remains fairly simple and lacks flexibility in scenarios with unsuccessful subtasks.

Therefore, we can improve our agent by implementing better skill calls, making it more robust to failures and incomplete subtasks. A promising solution would be an NN-based policy that processes all input data from all sensors to determine which skill to execute at the moment or whether a skill has been successfully completed, remains unfinished, or has failed.

Furthermore, our participation in the final challenge phase, where the top three teams presented their solutions, revealed that the top two teams both utilized heuristic agents with significantly modified main policies, resulting in substantial performance improvements. This provides important additional insight and suggests that there is considerable room for improvement in this agent component.

\subsubsection{Skill Training/Fine-Tuning}

In addition to enhancing skill calls through a more robust high-level policy, there is a need to improve each individual skill. For the majority of the challenge, we relied on baseline checkpoints (v1) provided within the initial HomeRobot repository. By leveraging additional resources, we successfully fine-tuned the place skill, resulting in overall performance improvements as previously demonstrated. Furthermore, after the challenge was over we tested new skill checkpoints (v2) that had been added to the repository. Incorporating these checkpoints into our agent and utilizing Ground-Truth segmentation led to an over 3\% improvement in overall success compared to previous results. However, we believe this is not the limit, and further advancements can be made in each skill.

\subsection{World Representation}
% Vladyslav
One of the key features of the OVMM task is that the environment remains the same across all subtasks. While this characteristic may not be crucial for tasks like \subtask{pick}/\subtask{gaze} and \subtask{place}, focusing solely on the object and receptacle within immediate sight, it becomes significantly relevant for navigation subtasks. During \subtask{navigation to object}, the lack of pre-existing data necessitates exploration of the environment. However, when navigating to the target receptacle, this exploration can be potentially enhanced by transferring or storing environment-related information obtained during first navigation skill.

One potential solution to address this challenge is employing a world representation of the environment. By remembering the locations of various objects, receptacles and the optimal paths to reach them, we can optimize exploration and enhance the agent's action selection. Additionally, dedicated data storage removes concerns about forgetting previously gathered information stored within RNN layers. Without such a world representation, the agent risks simply erasing previously encountered objects from its LSTM memory, leading to excessive and unnecessary environment exploration.

\section{Conclusions}
\label{section:conclusions}

We reported improvements of RL-based baseline for OVMM task. Our agent utilized improved perception, place skill, and high-level policy to obtain 5 times higher overall success rate (0.8\% to 5.2\%) and 2.5 times higher partial success rate (9.7\% to 25.7\%) on validation split, and 7 times higher overall success rate (0.4\% to 2.8\%) and 1.75 times higher partial success rate (10.9\% to 19.1\%) on Test Standard split of OVMM challenge dataset, scoring 3rd place in the simulation stage, and later in the real-world stage.

While achieving quite high ranking place, our approach does not solve the stated task, leaving much space for further improvements. Further work in the domain of semantic segmentation is necessary for achieving results in numerous tasks in EAI. Hopefully, the ideas presented in this report can build a foundation for future solution of the OVMM problem and similar problems in the field.

\bibliographystyle{abbrv}
\bibliography{main}

\end{document}